\documentclass{article} %
\usepackage{preprint,times}

\usepackage[utf8]{inputenc}
\usepackage[T1]{fontenc}
\usepackage{hyperref}
\usepackage{url}
\usepackage{booktabs}
\usepackage{multirow}
\usepackage{amsfonts}
\usepackage{amsmath}
\usepackage{graphicx}
\usepackage{algorithm}
\usepackage{algpseudocode}
\usepackage{xcolor}

\definecolor{nicered}{RGB}{196,30,58}
\definecolor{nicegreen}{RGB}{35,131,71}
\definecolor{niceyellow}{RGB}{181,136,6}
\definecolor{niceblue}{RGB}{28,97,178}

\title{Loss-Guided Tsetlin Machines: Making the Learning Objective Explicit}

\author{
    Mayur Shende$^{1}$ \quad Runar Helin$^{1}$ \quad Ole-Christoffer Granmo$^{1}$ \quad Vladimir I. Zadorozhny$^{2}$ \\
    $^{1}$University of Agder, Norway \quad $^{2}$University of Pittsburgh, USA
}

\preprintfinalcopy
\begin{document}

\maketitle

\begin{abstract}
    The Tsetlin Machine (TM) has no explicit notion of a loss function.
    Its learning signal is fixed to an implicit, margin-based update rule (MBUR), with no way to choose an objective function suited to a specific task.
    This paper introduces Loss-Guided TM, which replaces MBUR with an explicit and swappable loss function.
    In our proposed method, clause feedback and weight updates are driven by the loss, while keeping the Tsetlin Automaton and clause logic unchanged.
    The approach incorporates cross entropy (CE), symmetric cross entropy (SCE), and asymmetric loss (ASL).
    We evaluate Loss-Guided TM across six datasets that span multiclass, multilabel, and binary classification tasks.
    With a suitable loss, Loss-Guided TM matches or improves accuracy and F1 over MBUR on five of the six datasets.
    The gains are greatest for the imbalanced multilabel datasets, where ASL improves macro F1 by up to $9$ percentage points.
    With CE, Loss-Guided TM is also significantly better calibrated than MBUR on all six datasets, reducing calibration error by a factor of $3$ to $11$.
    A qualitative inspection shows that the clause-based global interpretability, examined at the per-class level on image and text data, is preserved when using the Loss-Guided mechanism.
    These results show that our explicit TM learning objective allows direct targeting of uncertainty calibration, data imbalance, and task structure, opening the TM to the broader literature on loss functions.
\end{abstract}

\section{Introduction}
\label{sec:intro}

The Tsetlin Machine (TM) is a logic-based alternative to neural networks, applied across various domains, such as image classification \citep{granmoConvolutionalTsetlinMachine2019, gronningsaeterOptimizedToolboxAdvanced2025}, text categorization and natural language understanding \citep{berge2019UsingTsetlinMachine, saha2022relationaltsetlinmachine, saha2023UsingTsetlinMachine, bhattarai2020MeasuringNoveltyNatural, gao2026llm}, regression \citep{darshanaabeyrathna2019regressionTsetlinmachine}, contextual bandits \citep{seraj2022TsetlinMachineSolving}, and board-game winner prediction \citep{giriLogicbasedAIInterpretable2022a}.
Its predictions are disjunctions of conjunctive clauses over binarized input literals, directly readable as propositional logic.
Together with a novel learning strategy (finding common denominators and reasoning by elimination), TMs become easier to interpret than neural networks \citep{yadav2022negation,shende2025Methodology, tarasyuk2026Learningdynamicspattern}.
Further work by \citet{kuruge2025ProbabilisticTsetlinMachine} and \citet{helin2025UncertaintyQuantificationTsetlin} quantifies the uncertainty of TM predictions.
TM learning has also been related to that of the perceptron \citep{sharmaEquivalenceWeightedTsetlin2022} and extended with regularization \citep{anjumNovelTsetlinMachine2024}.
The bit-level structure of the TM makes it efficient to run, with reported reductions in computation cost and memory footprint over comparable neural architectures \citep{kundu2026ComprehensiveReviewTsetlin}.
This combination of competitive accuracy, interpretability, and efficiency positions the TM as a practical, transparent alternative to deep learning \citep{lei2020ArithmeticLogicbased}.

A loss function defines what a model is actually being trained to do.
Different loss functions can produce different learned behavior even from the same architecture and the same data \citep{wasserman2004statistics}.
Cross entropy, the standard choice for classification, corresponds to the model's output representing a categorical or Bernoulli distribution over the labels, while mean squared error corresponds to a Gaussian distribution \citep{goodfellow2016deep}.
More broadly, loss functions in the literature are designed around the structure of a specific problem.
This includes, among many others \citep{wang2022loss}, losses built for extreme class imbalance \citep{lin2020focal}, for maximizing the margin between classes \citep{cortes1995support}, for region overlap in segmentation \citep{milletari2016vnet}, for matching entire distributions rather than individual predictions \citep{arjovsky2017wasserstein}, and for balancing reconstruction accuracy against a regularized latent representation~\citep{kingma2014autoencoding}.

However, TMs do not have an explicit loss function as a neural network does.
Instead, a TM uses a heuristic update probability~(Eq.~\ref{eq:classic-update-prob}), which is the normalized absolute error between the model's vote sum and the voting threshold~($T$).
This update mechanism, here termed margin-based update rule (MBUR), is hardwired directly into the TM's learning algorithm, with no principled bridge to the concept of loss functions.
The TM therefore cannot optimize its performance using any of the well-established objective functions, tailored to the various learning tasks.
This paper introduces a Loss-Guided learning mechanism for the TM that replaces MBUR's  heuristic update probability with an explicit loss function, while retaining the Tsetlin Automaton based logical rule learning.
We evaluate this mechanism using three different loss functions, cross entropy (CE), symmetric cross entropy (SCE), and asymmetric loss (ASL), assessing both predictive performance and interpretability against MBUR across six datasets.

\section{Background}
\label{sec:background}

TM is a machine learning model built from finite-state learning automata and propositional logic~\citep{granmo2018tsetlin}.
The Tsetlin Automaton (TA) is the model's basic learning unit, a finite-state machine that learns to select one of two actions, include or exclude, based on reward and penalty signals from its environment.
Each input feature is binarized and paired with its negation, forming the set of literals. Each literal is assigned a TA, which decides whether to include or exclude the literal.
Thus, a single propositional logic clause is defined by a dedicated team of TAs that operate on the literal set. In brief, the clause computes the conjunction $c_k = \bigwedge_{f \in L_k} f$ of the literals its TAs choose to include, where $L_k$ is the subset of literals included by clause $k$'s automata.
A clause is active, $c_k = 1$, when every literal in $L_k$ is true for the current input, and inactive, $c_k = 0$, otherwise.

The TM is initialized with the hyperparameters $C$ clauses, threshold $T$, specificity $s$, and negative-class sampling rate $q$.
Here, we are considering the weighted coalesced variant of the TM~\citep{glimsdal2021coalesced}.
Thus, each of the $C$ clauses is shared among the classes.
Each clause has a weight per class, $w_{ik}$ for clause $k$ and class $i$, whose sign gives the clause's polarity for that class.
Weights are initialized uniformly at random from $\{-1, 1\}$ and move in discrete $\pm 1$ steps during training~\citep{abeyrathna2020extending}.
The class sum~($v_i$) is the total vote of the model towards class $i$ and is calculated as:
\begin{equation}
    v_i = \sum_k w_{ik} \, c_k.
    \label{eq:classic-vote-sum}
\end{equation}
For multiclass classification, the predicted class is the one with the highest class sum.
For multilabel classification, a class is predicted whenever its class sum is positive.

During training, the probability of updating a clause~($P_i$) from feedback is given by:
\begin{equation}
    P_i = \frac{|y_i \cdot T - \operatorname{clip}(v_i, -T, T)|}{2T}, \qquad y_i \in \{-1, 1\}.
    \label{eq:classic-update-prob}
\end{equation}
When there are multiple classes, the probability of a class with label~$0$ giving feedback is further gated by probability $q / (M - 1)$, where $M$ is the number of classes.
Thus, $q$ controls the amount of negative feedback given by classes with label~$0$ and is one of the key parameters in multilabel classification.
A clause can receive three types of feedback, namely, \textit{Recognize}~(Type-Ia), \textit{Erase}~(Type-Ib), and \textit{Reject}~(Type-II).
For a class $i$ with label~$1$, the active positive polarity clauses receive Type-Ia, the inactive positive polarity clauses receive Type-Ib, and the active negative polarity clauses receive Type-II feedback. Similarly, for a class $i$ with label~$0$, the active positive polarity clauses receive Type-II feedback, inactive negative polarity clauses receive Type-Ib, and active negative polarity clauses receive Type-Ia feedback.
Each feedback type and how the literals are updated in each case is explained in detail in \citet{granmo2018tsetlin}.
In this paper, we refer to the TM trained under this margin-based update rule simply as MBUR.

\section{Methodology}

Given a sample $x$ and its labels $y$, each clause $k$ is evaluated against $x$, resulting in a clause output $c_k \in \{0, 1\}$.
In the proposed loss-guided scheme, the model output $\hat{y}_i$ for the class $i$ is given by,
\begin{equation}
    \hat{y}_i = A(z_i), \qquad \, z_i = \left(\frac{v_i}{C/2}\right)
    \label{eq:class-sum}
\end{equation}
where $A$ is an activation function, $v_i$ is the vote sum (Eq.~\ref{eq:classic-vote-sum}), and $z_i$ is the normalized vote sum.
The normalization makes $z_i$ independent of the number of clauses, preventing the activation from saturating for a large number of clauses $C$.
$C/2$ is used because the clauses are divided into sets of positive and negative polarity clauses, each with opposite learning objectives.
In the experiments, $A$ is the softmax function for multiclass classification and sigmoid otherwise.
Let $\ell(y, \hat{y})$ be the loss function. The loss $L = \ell(y, \hat{y})$ then quantifies the discrepancy between the predictions $\hat{y}$ of the model and the labels~$y$.

\paragraph{Clause updates}
We estimate the effect a clause has on the loss ($L$) by calculating $\Delta L_k$.
\begin{equation}
    \Delta L_k = L - L_{-k} , \qquad L_{-k} = \ell(y, \hat{y}_{-k})
    \label{eq:deltal}
\end{equation}
where $\hat{y}_{-k}$ is the model output obtained by inverting the clause output to $\bar{c}_k$, and $L_{-k}$ is the loss using this new output.
Thus, $\Delta L_k$ quantifies whether inverting clause $k$'s output reduces the loss.
We then evaluate the probability of updating a clause $k$, $P_k$, as
\begin{equation}
    P_k = 1 - e^{-\lambda |\Delta L_k|}
    \label{eq:update-prob}
\end{equation}
where $\lambda > 0$ is the feedback rate parameter. The function $1 - e^{-\lambda x}$ is the cumulative distribution function of the exponential distribution with rate $\lambda$, mapping the non-negative correction signal $|\Delta L_k|$ to a valid probability.
Unlike MBUR, where each class's vote sum is used to calculate update probability, here each clause only receives one update regardless of the number of classes.
The type of feedback a clause receives is determined by $\Delta L_k$ and $c_k$, as shown in Eq.~\ref{eq:feedback_selection}.

\begin{equation}
    \text{feedback}(c_k, \Delta L_k) =
    \begin{cases}
        \text{Type Ia} & \text{if } c_k = 1 \text{ and } \Delta L_k < 0 \\
        \text{Type Ib} & \text{if } c_k = 0 \text{ and } \Delta L_k > 0 \\
        \text{Type II} & \text{if } c_k = 1 \text{ and } \Delta L_k > 0 \\
        \text{None}    & \text{otherwise}
    \end{cases}
    \label{eq:feedback_selection}
\end{equation}

\paragraph{Weight Updates}
Clause weights $w_{ik} \in \mathbb{R}$ are initialized uniformly at random in $[-1, 1]$.
The weights are updated deterministically, and are decoupled from the stochastic TA feedback scheme.
The weight update follows the gradient of the loss with respect to each weight, obtained via the chain rule through $\hat{y}_i = A(z_i)$:
\begin{equation}
    \nabla w_{ik} = \frac{\partial \ell}{\partial w_{ik}} = \frac{\partial \ell}{\partial z_i} \cdot \frac{\partial z_i}{\partial w_{ik}} = \frac{c_k}{C/2} \, \frac{\partial \ell}{\partial z_i}
    \label{eq:weight-grad-derivation}
\end{equation}
Let $g_i = \partial \ell/\partial z_i$ denote the gradient of the loss with respect to the normalized vote.
For all the $\ell$ and $A$ pairs used in the experiments here, $g_i$ has a closed form.
Absorbing the constant factor $C/2$ into the learning rate $\eta$ gives the weight update,
\begin{equation}
    w_{ik} \leftarrow w_{ik} - \eta \, g_i \, c_k.
    \label{eq:weight-update}
\end{equation}
The clause feedback and the weight updates are thus both derived from the same explicit loss, and form a single learning mechanism rather than separate changes to MBUR.
Further experiments (Appendix~\ref{sec:appendix-ablation}) suggest that neither is effective without the other.

\paragraph{Hyperparameter changes}
The proposed mechanism does not depend on the \textit{threshold}~($T$).
Similarly, the hyperparameter $q$ is also no longer needed, since in the proposed approach, a clause receives feedback at most once per sample.
Instead, these are replaced by the \textit{learning rate} ($\eta$) and the \textit{feedback rate} ($\lambda$).

Algorithm~\ref{alg:training-epoch} summarizes one training epoch under the proposed mechanism.

\begin{algorithm}
    \caption{One training epoch}
    \label{alg:training-epoch}
    \begin{algorithmic}[1]
        \Require training set $\{(x_n, y_n)\}_{n=1}^N$, learning rate $\eta$, feedback rate $\lambda$
        \For{each training sample $(x, y)$}
        \For{each clause $k = 1, \dots, C$}
        \State Evaluate clause $k$ against $x$, giving clause output $c_k \in \{0, 1\}$
        \EndFor
        \For{each class $i$}
        \State $z_i \gets \dfrac{2}{C}\sum_k w_{ik} \, c_k$
        \State $\hat y_i \gets A(z_i)$ \Comment{Eq.~\ref{eq:class-sum}}
        \EndFor
        \State $L \gets \ell(y, \hat y)$
        \For{each clause $k = 1, \dots, C$}
        \For{each class $i$}
        \State $z'_i \gets z_i - \dfrac{2}{C}(c_k - \bar{c}_k)\, w_{ik}$ \Comment{normalized vote with inverted clause output $\bar{c}_k$}
        \State $\hat y'_i \gets A(z'_i)$
        \EndFor
        \State $\Delta L_k \gets L - \ell(y, \hat y')$ \Comment{Eq.~\ref{eq:deltal}}
        \State $P_k \gets 1 - e^{-\lambda |\Delta L_k|}$ \Comment{Eq.~\ref{eq:update-prob}}
        \If{$u \sim \text{Uniform}(0,1) < P_k$}
        \State Apply feedback$(c_k, \Delta L_k)$ to clause $k$  \Comment{Eq.~\ref{eq:feedback_selection}}
        \EndIf
        \For{each class $i = 1, \dots, M$}
        \State $g_i \gets \nabla_{z_i}\, L$
        \State $w_{ik} \gets w_{ik} - \eta \, g_i \, c_k$ \Comment{Eq.~\ref{eq:weight-update}}
        \EndFor
        \EndFor
        \EndFor
    \end{algorithmic}
\end{algorithm}

\section{Experimental Design}
\label{sec:experimental-design}

\subsection{Datasets}
\textbf{MNIST}~\citep{lecun1998gradient} is a standard image classification benchmark, containing 60,000 training and 10,000 test grayscale images across 10 digit classes.
\textbf{FashionMNIST}~\citep{xiao2017fashionmnist} is similar to MNIST, with the same train/test split sizes across 10 clothing-item classes.
\textbf{CelebA}~\citep{liu2015celeba} is a multilabel face-attribute classification dataset, containing 162,770 training and 19,867 test images, each labeled with 40 binary facial attributes.
This is a highly imbalanced dataset and contains $9.03$ labels per sample on average.
For this dataset, images are resized to $64\times64$ RGB and random horizontal flipping is applied during training.
Since the TM needs binarized inputs, we use 8-level thermometer encoding~\citep{buckman2018thermometer} on the pixel values, except for MNIST, which is thresholded using a static pixel intensity of 75.

\textbf{IMDB} is a binary movie review sentiment classification task, containing 25,000 training and 25,000 test examples.
The text is preprocessed by removing stopwords and encoding words into binary bag-of-words (unigram and bigram) features.
We then select the top 12,000 features using a chi-squared test.
\textbf{AG News}~\citep{zhang2015characterlevel} is a multiclass text classification task, containing 120,000 training and 7,600 test examples across 4 news-topic classes.
We follow preprocessing similar to IMDB, with an additional lemmatization step, using only unigram features, and selecting the top 2,000 features.
\textbf{GoEmotions}~\citep{demszky2020goemotions} is a multilabel emotion classification task, containing 43,400 training and 5,400 test examples across 28 emotion labels.
This is a highly imbalanced dataset, averaging 1.18 labels per sample.
Text is tokenized with a tweet-aware tokenizer that preserves case, collapses repeated characters, and properly handles emojis and unicode characters.
We then remove stopwords, lemmatize, replace URLs with a placeholder, and encode the top 10,000 features selected from unigram and bigram bag-of-words.

\subsection{Implementation and Evaluation Details}

MBUR and Loss-Guided TM are compared under matched capacity, meaning the clauses and specificity $s$ hyperparameters are kept the same across the two models.
All the models are trained for 100 epochs and use coalesced clauses.
We did not conduct an extensive hyperparameter grid search to optimize each condition.
Instead, hyperparameters for MBUR were selected through initial trials based on prior literature and understanding of $T$ and $q$.
For Loss-Guided TM, a similar approach was used to select the learning rate $\eta$ and feedback rate $\lambda$.
Table~\ref{tab:hyperparameters} lists the selected hyperparameters used for each experiment.

We use three different loss functions across datasets to demonstrate the use of different losses according to the dataset.
Cross entropy (CE) is the standard classification loss, measuring the divergence between the predicted and true class distributions.
Symmetric cross entropy (SCE) \citep{wang2019symmetric} adds a reverse cross entropy term that stays bounded even when the model is confidently wrong, improving robustness to label noise.
Asymmetric loss (ASL) \citep{ridnik2021asymmetric} applies separate down-weighting factors to positive and negative predictions along with a hard threshold on the gradient for confident negatives.
This is designed to counteract the positive-negative imbalance inherent to multilabel classification.
For brevity, we refer to the Loss-Guided TM trained with CE, SCE, or ASL simply by the name of the loss.

\begin{table}[htbp]
    \caption{Hyperparameters used for each dataset and condition. $T$ and $q$ are used by MBUR, while $\eta$ (learning rate) and $\lambda$ (feedback rate) are used by Loss-Guided TM.}
    \label{tab:hyperparameters}
    \begin{center}
        \resizebox{\linewidth}{!}{%
            \begin{tabular}{lccccccccc}
                \toprule
                \textbf{Dataset} & \textbf{$C$} & \textbf{$s$} & \textbf{$T$} & \textbf{$q$} & \textbf{$p\times p$} & \textbf{$\ell$} & \textbf{$\eta$} & \textbf{$\lambda$} \\
                \midrule
                MNIST & 1000 & 10.0 & 2000 & 1.0 & $10\times10$ & CE & 1.0 & 1.0 \\
                \midrule
                FashionMNIST & 16000 & 10.0 & 20000 & 1.0 & $3\times3$ & CE & 0.5 & 3.0 \\
                \midrule
                IMDB & 10000 & 1.1 & 10000 & 1.0 & - & CE & 0.01 & 1.0 \\
                \midrule
                \multirow{2}{*}{AG News} & \multirow{2}{*}{8000} & \multirow{2}{*}{2.0} & \multirow{2}{*}{40000} & \multirow{2}{*}{1.0} & \multirow{2}{*}{-} & CE & 0.01 & 1.0 \\
                & & & & & & SCE ($\alpha{=}1.0$, $\beta{=}1.0$) & 0.005 & 1.0 \\
                \midrule
                \multirow{2}{*}{GoEmotions} & \multirow{2}{*}{22400} & \multirow{2}{*}{2.0} & \multirow{2}{*}{25000} & \multirow{2}{*}{1.5} & \multirow{2}{*}{-} & CE & 0.5 & 4.0 \\
                & & & & & & ASL ($\gamma_+{=}0.0$, $\gamma_-{=}4.0$, clip${=}0.01$) & 0.05 & 6.0 \\
                \midrule
                \multirow{2}{*}{CelebA} & \multirow{2}{*}{80000} & \multirow{2}{*}{30.0} & \multirow{2}{*}{80000} & \multirow{2}{*}{10.0} & \multirow{2}{*}{$3\times3$} & CE & 1.0 & 25.0 \\
                & & & & & & ASL ($\gamma_+{=}2.0$, $\gamma_-{=}4.0$, clip${=}0.01$) & 1.0 & 25.0 \\
                \bottomrule
            \end{tabular}%
        }
    \end{center}
\end{table}

For multiclass classification tasks, accuracy, macro F1, macro ROC-AUC, macro PR-AUC, and macro ECE (expected calibration error) at the final epoch are reported.
ECE measures how well a model's predicted confidence matches how often it is actually correct.
It is relevant whenever a confidence score is used for more than just picking the top prediction, for example flagging uncertain cases for review.
For MBUR, the vote sums are converted to probabilities using the probability score of \citet{helin2025UncertaintyQuantificationTsetlin},
\begin{equation}
    \hat{y}_i = \frac{1}{2}\left(1 + \frac{\operatorname{clip}(v_i, -T, T)}{T}\right),
    \label{eq:mbur-prob}
\end{equation}
which are used for ECE, ROC-AUC and PR-AUC.

For multilabel classification, macro F1, weighted F1, ROC-AUC, PR-AUC, and macro ECE are reported instead.
Accuracy is uninformative for multilabel classification and is therefore omitted.
ROC-AUC and PR-AUC are threshold-independent ranking metrics, better suited to imbalanced and multilabel settings than a fixed-threshold metric like F1.
PR-AUC is the more informative of the two under heavy label imbalance, since ROC-AUC can stay high even when a classifier performs poorly on the rare positive label.

Each dataset experiment is repeated with 10 seeds for each of the conditions.
Differences in the metrics are tested with the paired Wilcoxon signed-rank test \citep{wilcoxon1945individual}.
Each dataset involves multiple pairwise comparisons.
Thus, raw $p$-values are adjusted with the Holm-Bonferroni correction \citep{holm1979simple} to control the family-wise error rate.
Effect size is measured alongside every $p$-value as paired Cohen's $d_z$ \citep{lakens2013calculating}, to estimate the magnitude of the difference in the results.
In the tables~\ref{tab:multiclass-results} and \ref{tab:multilabel-binary-results}, $^{\dagger}$ denotes significance ($p<0.05$) relative to MBUR.

\section{Results}

Table~\ref{tab:multiclass-results} shows the test set metrics for the multiclass classification datasets.
Across all the datasets tested here, the Loss-Guided TM achieves substantially better calibration, with a large and consistent effect size.
Accuracy and F1 are comparable or better than MBUR.
The largest gain in accuracy was seen on the IMDB dataset, with a very large effect size ($d_z \approx 18.7$), reflecting highly consistent and low-variance improvement across seeds.
On AG News, SCE achieves the highest accuracy of all conditions, at the cost of a smaller calibration gain compared to CE.

\begin{table}[htbp]
    \caption{Multiclass classification test metrics (mean $\pm$ std over 10 seeds).}
    \label{tab:multiclass-results}
    \begin{center}
        \resizebox{\linewidth}{!}{%
            \begin{tabular}{llccccc}
                \toprule
                \textbf{Dataset} & \textbf{Condition} & \textbf{Accuracy (\%)} & \textbf{F1 (\%)} & \textbf{ROC-AUC (\%)} & \textbf{PR-AUC (\%)} & \textbf{ECE} \\
                \midrule
                \multirow{2}{*}{MNIST} & MBUR & \textbf{98.75 $\pm$ 0.11} & \textbf{98.74 $\pm$ 0.11} & 99.92 $\pm$ 0.01 & 99.46 $\pm$ 0.06 & 0.0056 $\pm$ 0.0002 \\
                & CE & 98.55 $\pm$ 0.13 & 98.54 $\pm$ 0.13 & \textbf{99.98 $\pm$ 0.00}$^{\dagger}$ & \textbf{99.87 $\pm$ 0.02}$^{\dagger}$ & \textbf{0.0016 $\pm$ 0.0001}$^{\dagger}$ \\
                \midrule
                \multirow{2}{*}{FashionMNIST} & MBUR & 90.32 $\pm$ 0.13 & 90.28 $\pm$ 0.15 & 98.85 $\pm$ 0.04 & 92.49 $\pm$ 0.07 & 0.0263 $\pm$ 0.0009 \\
                & CE & \textbf{90.49 $\pm$ 0.18}$^{\dagger}$ & \textbf{90.46 $\pm$ 0.16}$^{\dagger}$ & \textbf{99.31 $\pm$ 0.01}$^{\dagger}$ & \textbf{95.37 $\pm$ 0.08}$^{\dagger}$ & \textbf{0.0047 $\pm$ 0.0003}$^{\dagger}$ \\
                \midrule
                \multirow{2}{*}{IMDB} & MBUR & 85.52 $\pm$ 0.12 & 85.52 $\pm$ 0.12 & 90.20 $\pm$ 0.10 & 87.37 $\pm$ 0.23 & 0.0934 $\pm$ 0.0011 \\
                & CE & \textbf{88.61 $\pm$ 0.08}$^{\dagger}$ & \textbf{88.61 $\pm$ 0.08}$^{\dagger}$ & \textbf{95.14 $\pm$ 0.04}$^{\dagger}$ & \textbf{94.96 $\pm$ 0.04}$^{\dagger}$ & \textbf{0.0142 $\pm$ 0.0012}$^{\dagger}$ \\
                \midrule
                \multirow{3}{*}{AG News} & MBUR & 88.94 $\pm$ 0.24 & 88.91 $\pm$ 0.24 & 96.60 $\pm$ 0.09 & 90.48 $\pm$ 0.26 & 0.0648 $\pm$ 0.0039 \\
                & CE & 89.74 $\pm$ 0.15$^{\dagger}$ & 89.73 $\pm$ 0.15$^{\dagger}$ & \textbf{97.71 $\pm$ 0.02}$^{\dagger}$ & \textbf{94.32 $\pm$ 0.05}$^{\dagger}$ & \textbf{0.0109 $\pm$ 0.0006}$^{\dagger}$ \\
                & SCE & \textbf{90.38 $\pm$ 0.13}$^{\dagger}$ & \textbf{90.36 $\pm$ 0.13}$^{\dagger}$ & 97.50 $\pm$ 0.02$^{\dagger}$ & 93.98 $\pm$ 0.06$^{\dagger}$ & 0.0265 $\pm$ 0.0006$^{\dagger}$ \\
                \bottomrule
            \end{tabular}%
        }
    \end{center}
\end{table}

Table~\ref{tab:multilabel-binary-results} shows the results for multilabel classification datasets.
ASL achieves significantly better F1 scores (macro and weighted) on both datasets, with large effect sizes.
However, ASL also produces the least calibrated models, showing calibration error higher than MBUR.
Similar to the multiclass results, the CE condition achieved the lowest calibration error on both datasets.
For CelebA, the CE condition also achieved slightly better PR and ROC AUC scores, alongside weighted F1 comparable to ASL.
However, ASL achieves substantially higher macro F1.
This reflects the design of ASL, which counteracts multilabel imbalance and helps the model learn rare classes.

\begin{table}[htbp]
    \caption{Multilabel classification results on the test set (mean $\pm$ std over 10 seeds).}
    \label{tab:multilabel-binary-results}
    \begin{center}
        \resizebox{\linewidth}{!}{%
            \begin{tabular}{llccccc}
                \toprule
                \textbf{Dataset} & \textbf{Condition} & \textbf{F1 (macro, \%)} & \textbf{F1 (weighted, \%)} & \textbf{PR-AUC (\%)} & \textbf{ROC-AUC (\%)} & \textbf{ECE} \\
                \midrule
                \multirow{3}{*}{GoEmotions} & MBUR & 28.09 $\pm$ 1.08 & 38.03 $\pm$ 0.63 & 28.51 $\pm$ 0.60 & 80.55 $\pm$ 0.28 & 0.1601 $\pm$ 0.0076 \\
                & CE & 26.38 $\pm$ 0.37 & 40.42 $\pm$ 0.25$^{\dagger}$ & 30.69 $\pm$ 0.22$^{\dagger}$ & 81.60 $\pm$ 0.16$^{\dagger}$ & \textbf{0.0142 $\pm$ 0.0005}$^{\dagger}$ \\
                & ASL & \textbf{37.06 $\pm$ 0.28}$^{\dagger}$ & \textbf{50.44 $\pm$ 0.18}$^{\dagger}$ & \textbf{34.46 $\pm$ 0.15}$^{\dagger}$ & \textbf{82.45 $\pm$ 0.08}$^{\dagger}$ & 0.2396 $\pm$ 0.0008 \\
                \midrule
                \multirow{3}{*}{CelebA} & MBUR & 55.21 $\pm$ 0.67 & 67.99 $\pm$ 0.51 & 57.82 $\pm$ 0.16 & 86.08 $\pm$ 0.18 & 0.1457 $\pm$ 0.0150 \\
                & CE & 58.47 $\pm$ 0.50$^{\dagger}$ & 71.12 $\pm$ 0.25$^{\dagger}$ & \textbf{67.04 $\pm$ 0.10}$^{\dagger}$ & \textbf{89.60 $\pm$ 0.04}$^{\dagger}$ & \textbf{0.0299 $\pm$ 0.0021}$^{\dagger}$ \\
                & ASL & \textbf{63.86 $\pm$ 0.11}$^{\dagger}$ & \textbf{73.75 $\pm$ 0.09}$^{\dagger}$ & 66.73 $\pm$ 0.08$^{\dagger}$ & 89.44 $\pm$ 0.05$^{\dagger}$ & 0.1802 $\pm$ 0.0009 \\
                \bottomrule
            \end{tabular}%
        }
    \end{center}
\end{table}

Averaged over the ten rarest attributes, ASL's mean F1 gain over MBUR is +15.8 percentage points, compared to only +5.8 points over the ten most common attributes, a negative correlation between attribute frequency and gain, while CE shows no such correlation.
The effect is clearest on the rarest attributes.
On Bald (2.3\% of samples), ASL raises F1 from $30.0\pm8.6\%$ to $57.4\pm0.7\%$, and on Blurry (5.1\% of samples), from $14.0\pm4.4\%$ to $38.3\pm1.1\%$, in both cases also sharply reducing the across-seed variance.
CE does not share this pattern.
On Mustache (4.1\% of samples), CE lowers F1 relative to MBUR ($30.6\%\to27.6\%$), while ASL still improves it ($42.4\%$).
On the most common attributes the gap narrows sharply, e.g.\ No\_Beard (83.4\% of samples) improves only from $91.5\pm0.4\%$ to $94.4\pm0.1\%$.

On GoEmotions, classes grief (0.18\% of samples) and relief (0.35\% of samples) score close to zero F1 across all three conditions.
Where MBUR does show a non-zero mean, for example $5.1\pm10.8\%$ on grief, the variance is quite large.
This reflects the stochastic nature of TM learning, and indicates that this class in particular is not reliably learned.
ASL's F1 gain over MBUR grows with class frequency instead of shrinking, the opposite direction from CelebA.
On more frequent classes this gain is large, for example on amusement (5.4\% of samples), F1 rises from $47.4\pm3.8\%$ to $77.5\pm0.7\%$, and on gratitude (6.1\% of samples), from $66.0\pm4.4\%$ to $86.8\pm1.1\%$.
On some of the rarer classes ASL falls behind MBUR instead, for example on embarrassment (0.70\% of samples), F1 falls from $25.2\pm4.3\%$ to $7.7\pm2.6\%$.
The CE condition had the worst macro F1 of the three, since it was particularly difficult for it to learn some of the rarer classes.
On disapproval (4.7\% of samples), CE lowers F1 from $17.1\pm1.2\%$ to $10.2\pm1.5\%$, while ASL raises it to $23.1\pm1.2\%$.

\begin{figure}[htbp]
    \begin{center}
        \includegraphics[width=0.9\linewidth]{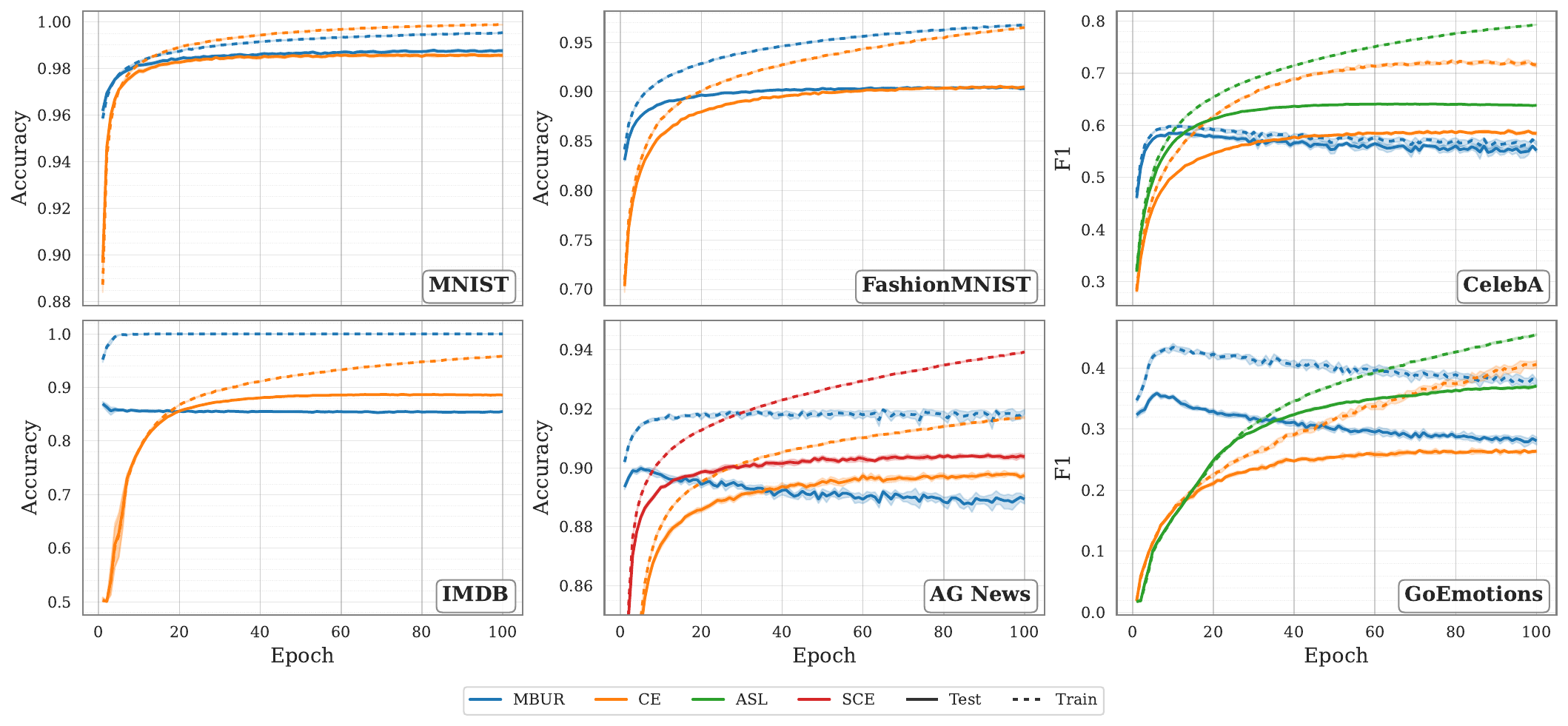}
    \end{center}
    \caption{Training and test curves over 100 epochs for all six datasets. Plots show accuracy for MNIST, FashionMNIST, IMDB, and AG News, and macro F1 for CelebA and GoEmotions. Solid curves denote test performance and dashed curves denote training performance for MBUR and Loss-Guided variants.}
    \label{fig:f1}
\end{figure}

Figure~\ref{fig:f1} shows the train and test curves over 100 training epochs for accuracy (for multiclass) or macro F1 (for multilabel) metrics.
The Loss-Guided models show a steadier increase and more stable learning curves compared to MBUR.
MBUR overfits heavily on the IMDB dataset, where the train accuracy reaches $100\%$ within the first few epochs.
In contrast, the Loss-Guided model shows a slower learning curve and lower degree of overfitting, while achieving higher test accuracy.
On AG News, GoEmotions, and CelebA, MBUR's performance peaks within the first few epochs, followed by a decline in both training and test performance. In contrast, Loss-Guided TM improves over more epochs, reaching higher peak test performance before showing signs of overfitting.

\subsection{Interpretability}
\label{sec:interpretability}

Beyond the predictive metrics reported above, we examine the global interpretations produced by the models.
Local interpretability highlights the parts of the input responsible for a given prediction, computed using the WAC score \citep{clarke2026TsetlinMachineThird}, while global interpretability aggregates all the learned patterns for a particular class.
Global interpretability therefore allows us to analyze and compare the combined contribution of all the clauses in the model.
We present this comparison for the CelebA and IMDB datasets.
Due to the stochastic nature of TM learning, each seed can produce a different set of clauses and slightly different interpretations.
Thus, we compare the interpretations averaged over the $10$ seeds.

\begin{figure}[htbp]
    \begin{center}
        \includegraphics[width=0.9\linewidth]{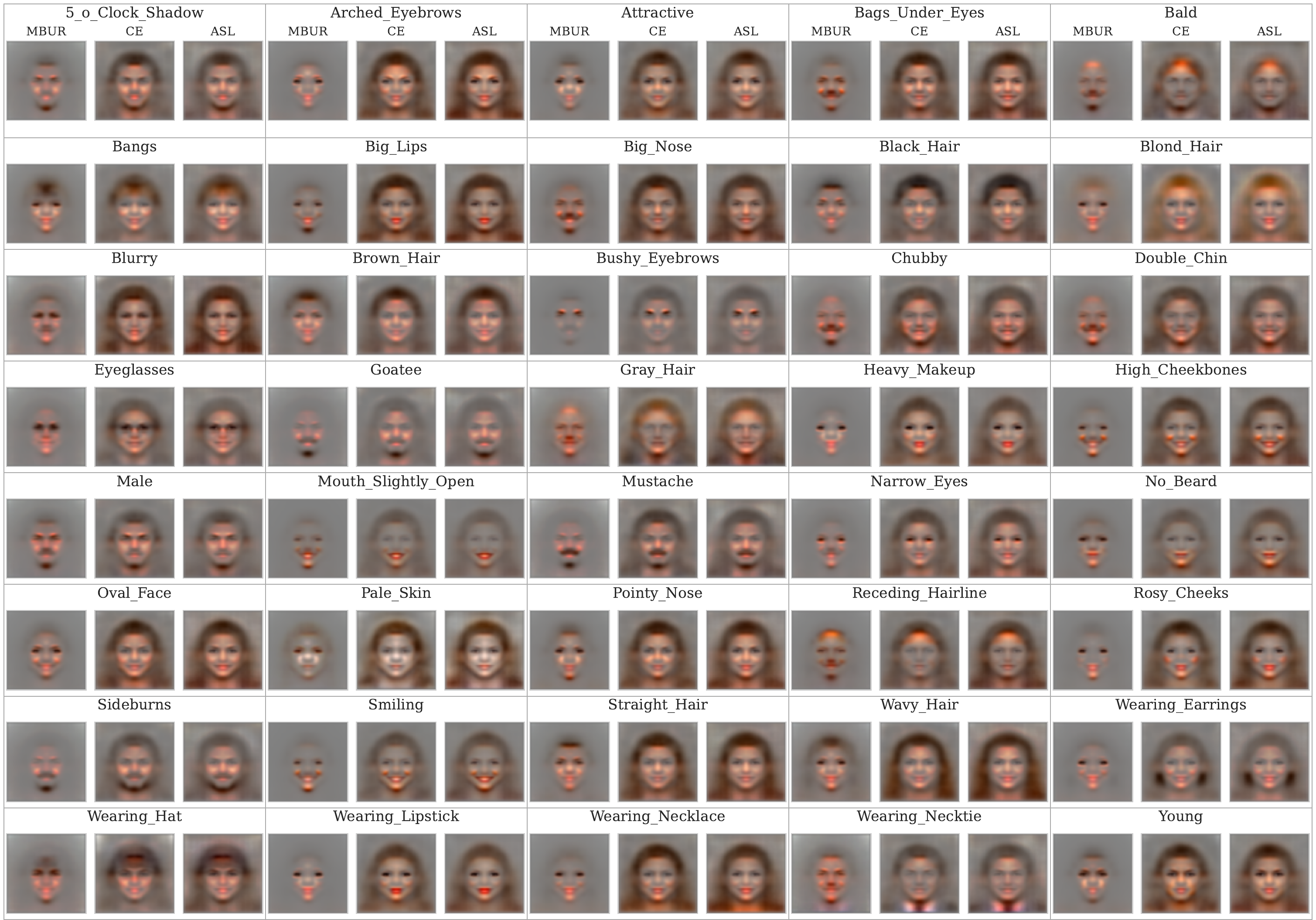}
    \end{center}
    \caption{Per-class global interpretations for all 40 CelebA facial attributes, comparing MBUR, Loss-Guided CE, and Loss-Guided ASL averaged over 10 seeds.}
    \label{fig:celeba-wic}
\end{figure}

For CelebA, we follow the global interpretability procedure for convolutional TM described by \citet{shende2025Methodology}.
Figure~\ref{fig:celeba-wic} shows these global interpretations for the 40 CelebA attributes.
Non-gray regions indicate pixels the model considers important, and gray regions indicate don't-care or unimportant regions.
CE and ASL produce visually similar interpretations across all 40 attributes.
For several attributes, MBUR already focuses on the correct patterns with reasonably strong importance, for example Bushy\_Eyebrows, Wearing\_Lipstick, and Smiling.
These are typically the classes for which MBUR has fairly high F1 score as well.
However, it fails to capture the correct pattern for several other classes.
On Wearing\_Earrings, MBUR shows no class-specific patterns.
On Wearing\_Necktie, MBUR focuses on the face rather than the relevant region.
On Blond\_Hair, Wearing\_Hat, and Wearing\_Necklace, MBUR gives very low importance to the relevant regions.
In all these cases, the Loss-Guided models produce better interpretations, and clearly capture more patterns for the classes.
MBUR also fails to distinguish between related attributes that Loss-Guided separates correctly.
For instance, MBUR shows nearly the same pattern for Mustache, Sideburns, and Goatee classes, while Loss-Guided models capture distinct patterns for the three attributes.
The same holds for Big\_Nose and Chubby, where MBUR's interpretations are close to identical, while CE and ASL produce clearly different patterns for the two.
More generally, Loss-Guided's interpretations cover a noticeably larger part of the face, whereas MBUR remains sparse and concentrated on a few small regions.

\begin{figure}[htbp]
    \begin{center}
        \includegraphics[width=0.9\linewidth]{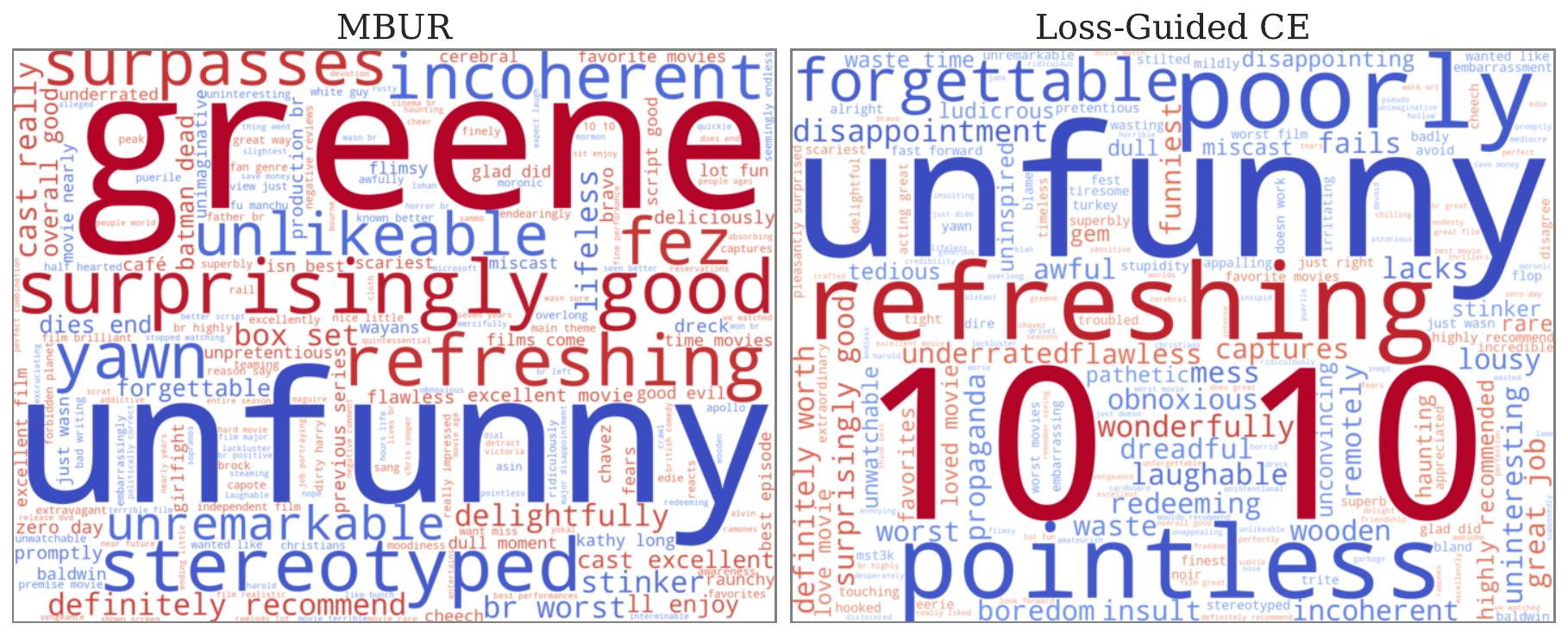}
    \end{center}
    \caption{Per-class global interpretation for the IMDB dataset under MBUR and Loss-Guided CE. Words in red indicate positive word importance, while blue indicates negative importance. The relative magnitude of the importance is shown using the size of the word and the intensity of the color.}
    \label{fig:imdb-wic-wordcloud}
\end{figure}

For IMDB, we calculate the interpretability using a procedure similar to that for the image dataset. However, here we additionally include the negative polarity clauses in the final importance score.
For each class, we take the weighted sum of positive-polarity clauses, which gives the combined contribution for the positive-polarity clauses.
Since the clauses contain non-negated and negated literals, we take a difference between the two parts to produce an effective per-feature contribution.
We repeat the same procedure for negative-polarity clauses to obtain their per-feature contribution.
The difference of the two polarity contributions, followed by the asymmetric normalization procedure of \citet{shende2025Methodology} gives the combined per-feature contribution of all the clauses for a class.
These are visualized as word clouds per class in Figure~\ref{fig:imdb-wic-wordcloud}.
Each class's own topic vocabulary emerges as positive evidence (shown here as red), and other classes' vocabulary as negative evidence (shown here as blue), consistently across the conditions.

Generic movie-review vocabulary with no inherent sentiment receives near-zero importance under both conditions, for example "story" (0.0033 vs 0.0024), "style" (0.0033 vs 0.0043), and "watching" (-0.0048 vs 0.0052), given as Loss-Guided vs MBUR.
A few words also receive near-zero importance despite sounding sentiment-laden, but these are simply rare in the training data.
"Stupid ending" and "absolutely horrendous" appear in only 6 and 7 training reviews respectively. These occurrences are too few for either model to learn a reliable pattern from.
The two conditions differ most on the words attributed with a high importance.
Loss-Guided gives canonical sentiment words a higher importance than MBUR, for both polarities, for example "worst" (-0.71 vs -0.35), "awful" (-0.70 vs -0.24), "10 10" (1.00 vs 0.66), and "wonderfully" (0.84 vs 0.49).
Loss-Guided also gives sentiment-irrelevant proper nouns a lower importance than MBUR, for example "greene" (0.46 vs 1.00), "fez" (0.28 vs 0.90), "wayans" (-0.43 vs -0.74), and "kathy long" (-0.30 vs -0.76).

\section{Discussion}
Loss-Guided TM's advantage over MBUR is not confined to a single metric. It holds across performance, calibration, and training stability, with the choice of loss determining where the largest gain falls.
With a suitable loss, Loss-Guided TM matches or improves accuracy and F1 over MBUR on five of the six datasets, and with CE it is better calibrated on all six.
This advantage persists when the same post-hoc calibration is applied to both models (Appendix~\ref{sec:appendix-posthoc}).
CE gives the best calibration but is not designed for imbalanced, multilabel tasks, while ASL trades calibration for performance in that setting.
This shows different losses are better suited for different tasks.

Because the Loss-Guided mechanism updates the TAs using the established feedback mechanism, the model still learns Boolean propositional rules over the features, preserving interpretability by construction.
The classes for which Loss-Guided models perform better, also produces better interpretations, as seen for the CelebA dataset.  
MBUR interpretations are weakest on the rare attributes where its F1 is lowest, whereas Loss-Guided TM concentrates its largest F1 gains on those same rare classes, producing clearer interpretations.

Loss-Guided learning is also more stable than MBUR (Figure~\ref{fig:f1}).
MBUR's early-peak-then-decline pattern on GoEmotions, CelebA, and AG News is different from overfitting, since train and test decline together, indicating degradation of the model itself.
In our limited tuning for the MBUR model, the same pattern was observed with several tested configurations, mostly in the case of multilabel datasets.
Loss-Guided models showed no such decline on any of the datasets, staying stable or continuing to improve for the rest of training.

Hyperparameters were fixed to one operating point per dataset, with the shared hyperparameters identical between MBUR and every Loss-Guided variant to compare the learning mechanisms directly.
These results describe a common point of comparison, not the best achievable configuration for either mechanism.
A proper hyperparameter search to identify optimal configurations is left for future studies.

Making the TM's objective explicit opens access to a much larger design space than the three losses tested here.
Loss functions are a decades-deep, mature area of machine learning research, developed for regression, imbalance, robustness, and structured prediction, and MBUR's fixed feedback rule gave the TM no way to draw on any of it.
The same substitution mechanism should also extend beyond a single loss term to regularizers, penalties added to shape what a model learns rather than just how it is scored.
How far this substitution mechanism extends, and what it makes possible beyond the classification tasks tested here, needs further research.

\section{Conclusion}

This paper presented a replacement of MBUR, the Tsetlin Machine's fixed, implicit feedback rule, with an explicit, swappable loss function, keeping the underlying Tsetlin Automata and clause logic unchanged.
Across six datasets spanning multiclass, multilabel, and binary classification, Loss-Guided TM with a suitable loss matches or improves F1 and accuracy over MBUR on five of the six datasets, and with CE achieves better calibration on all six.
Furthermore, specific losses align with specific task structures, as demonstrated by ASL achieving substantially higher performance on imbalanced multilabel datasets.
The Loss-Guided mechanism does not adversely affect the clause-based interpretability of the TM, and on some datasets produces clearer interpretations than MBUR.
By making the learning objective explicit, Loss-Guided TM opens the Tsetlin Machine to the broader loss function literature, developed over decades for calibration, imbalance, and task structure, giving a direct way to target these properties in future work.

\bibliography{references}
\bibliographystyle{preprint}

\appendix
\section{Appendix}

\subsection{MNIST and FashionMNIST}

\begin{figure}[htbp]
    \begin{center}
        \includegraphics[width=\linewidth]{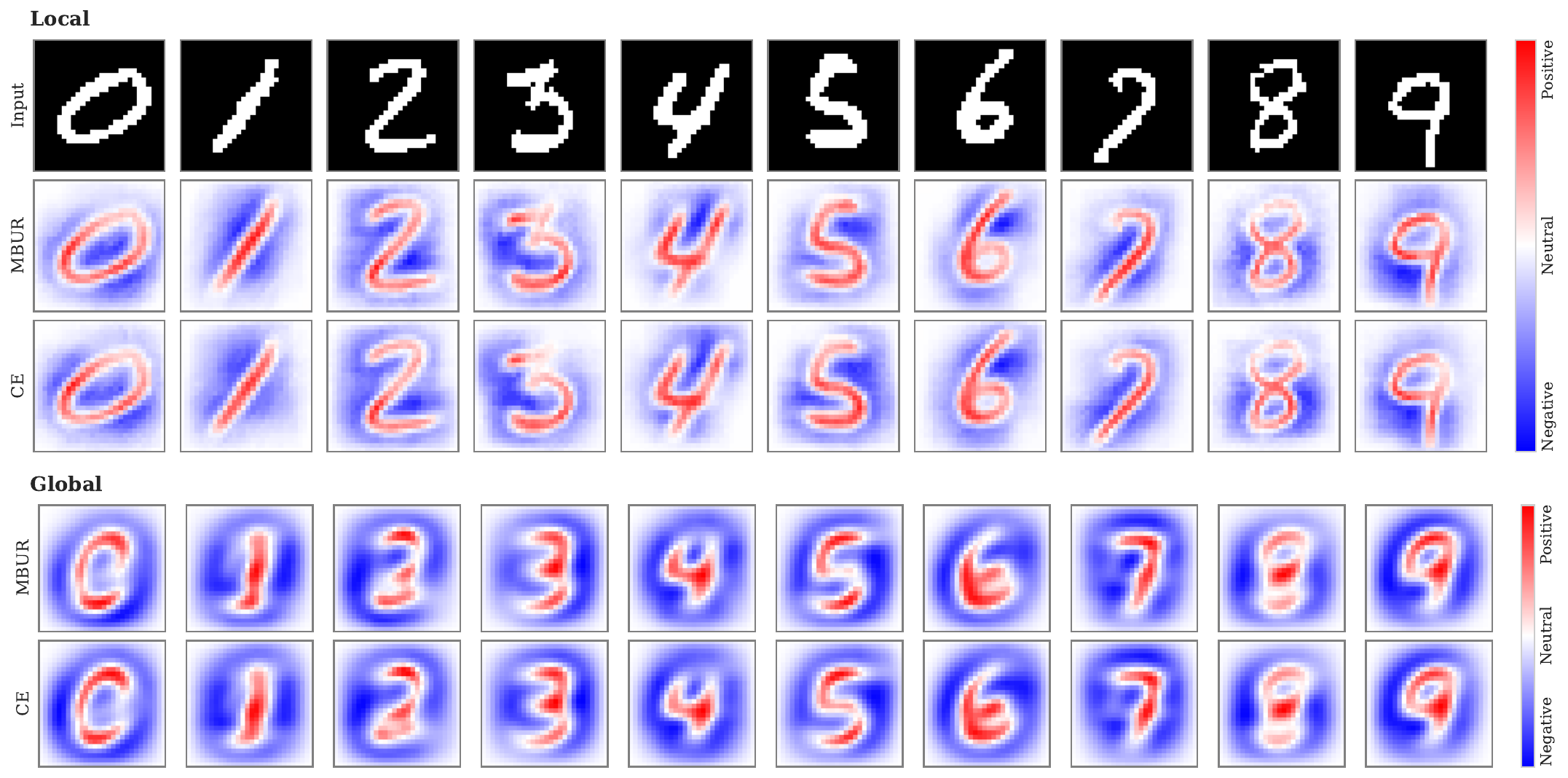}
    \end{center}
    \caption{The local and global interpretation for the MNIST dataset. Red indicates positive feature importance, while blue indicates negative importance. The color intensity indicates the relative magnitude of the importance. The white region ($\approx$ zero importance) indicates \textit{don't care} features, which do not affect the predictions of the model.}
    \label{fig:mnist-interpretation}
\end{figure}

\begin{figure}[htbp]
    \begin{center}
        \includegraphics[width=\linewidth]{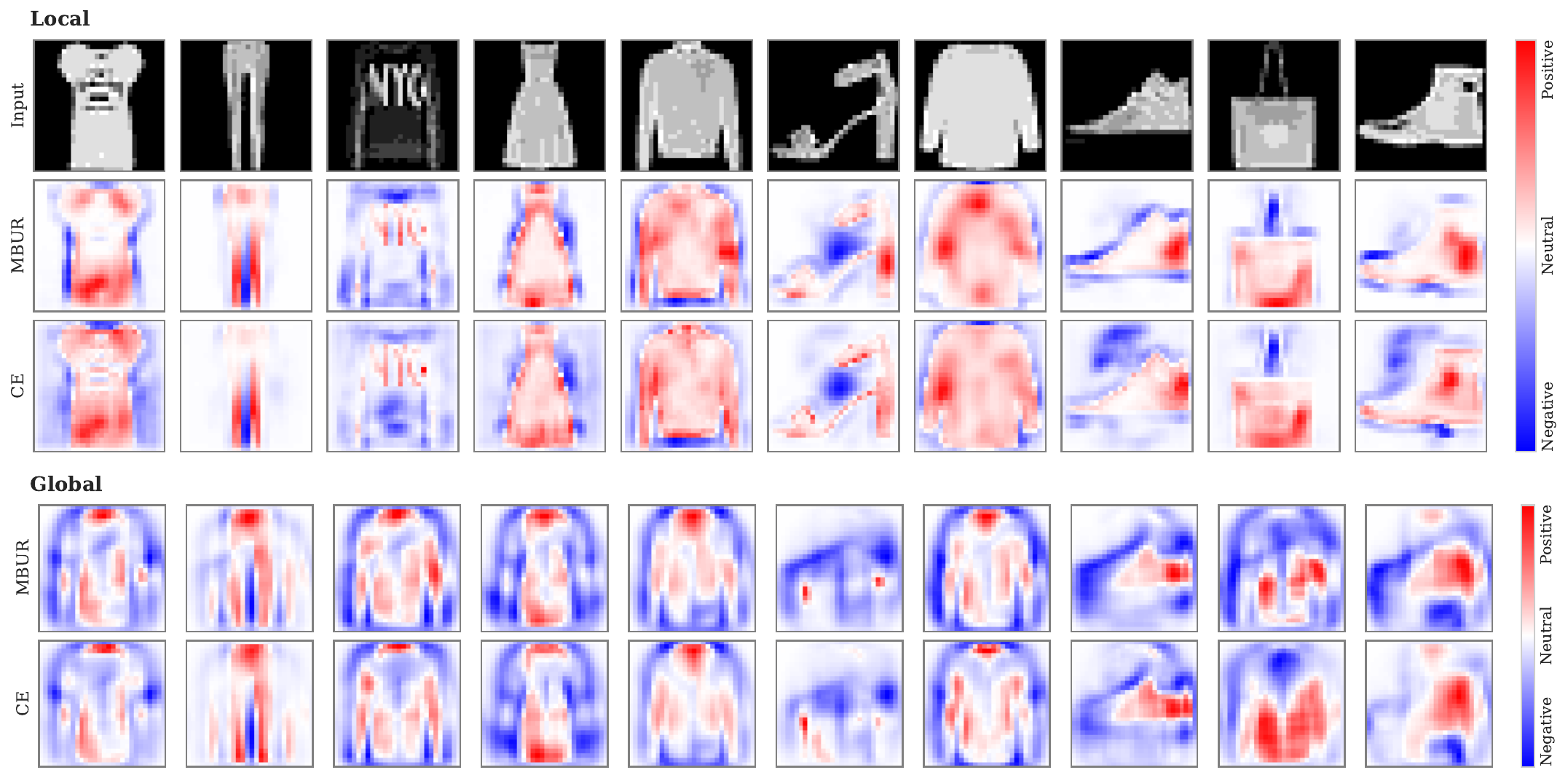}
    \end{center}
    \caption{The local and global interpretation for the FashionMNIST dataset. Red indicates positive feature importance, while blue indicates negative importance. The color intensity indicates the relative magnitude of the importance.}
    \label{fig:fmnist-interpret}
\end{figure}

For MNIST and FashionMNIST, we compare the global interpretation for both conditions, following the same convolutional TM procedure of \citet{shende2025Methodology} used for CelebA.
We additionally include the local interpretation for a few random samples from the test set.
Figures~\ref{fig:mnist-interpretation} and \ref{fig:fmnist-interpret} show the local and global interpretations for the two datasets, averaged over 10 seeds.
MBUR and CE are visually indistinguishable in both.

\subsection{IMDB}

\begin{figure}[htbp]
    \begin{center}
        \includegraphics[width=\linewidth]{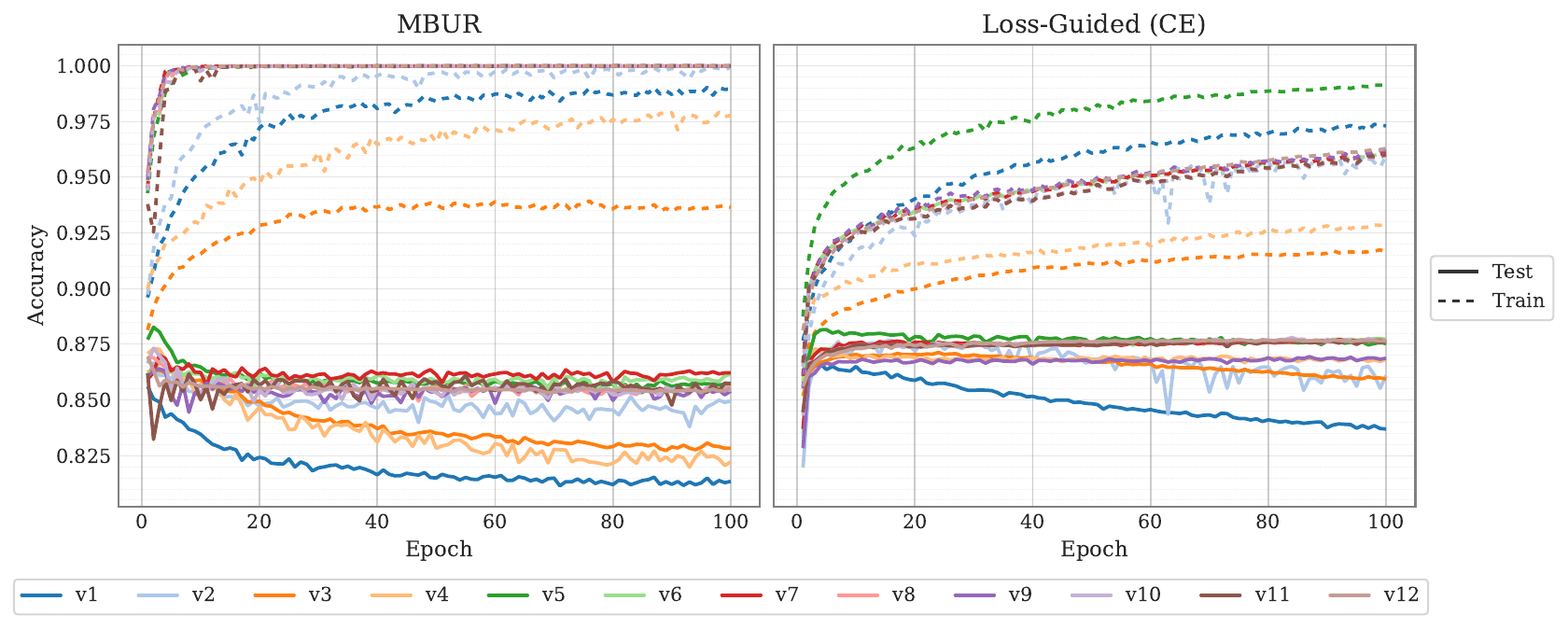}
    \end{center}
    \caption{Training and test accuracy curves for all twelve IMDB sweep configurations (Table~\ref{tab:imdb-sweep}), using MBUR (left) and Loss-Guided CE (right). Solid curves denote test accuracy, dashed curves denote train accuracy.}
    \label{fig:imdb-sweep}
\end{figure}

IMDB showed the largest gap between MBUR and Loss-Guided, both in accuracy and in overfitting behavior.
Before settling on the reported configuration, we swept clause count from 500 to 20,000, with the threshold $T$ and specificity $s$ scaled accordingly, across twelve trial configurations (Table~\ref{tab:imdb-sweep}).
MBUR overfit under every evaluated configuration, with train accuracy reaching 94\% to 100\% while test accuracy plateaued between 81\% and 86\%.
CE showed a substantially smaller train-test gap under the same sweep, with test accuracy consistently higher than MBUR's across every configuration tried (Figure~\ref{fig:imdb-sweep}).
IMDB was the only dataset that received a more extensive hyperparameter search, due to the observed learning behavior of MBUR. 

\begin{table}[htbp]
    \caption{Hyperparameters for the twelve IMDB sweep configurations (v1--v12). The Final row is the configuration reported in the main text (Table~\ref{tab:hyperparameters}), using the MBUR runs of v12.}
    \label{tab:imdb-sweep}
    \begin{center}
        \begin{tabular}{lccccccc}
            \toprule
            & & & & & & \multicolumn{2}{c}{\textbf{Test Accuracy}} \\
            \cmidrule(l){7-8}
            \textbf{Config} & \textbf{Clauses} & \textbf{$s$} & \textbf{$T$} & \textbf{$\eta$} & \textbf{Features} & MBUR & Loss-Guided CE \\
            \midrule
            v1 & 2000 & 2.0 & 5000 & 0.1 & 5000 & 0.813 & 0.837 \\
            v2 & 1000 & 2.0 & 5000 & 0.01 & 15000 & 0.849 & 0.861 \\
            v3 & 500 & 2.0 & 5000 & 0.01 & 2000 & 0.828 & 0.859 \\
            v4 & 500 & 1.1 & 5000 & 0.01 & 2000 & 0.822 & 0.868 \\
            v5 & 10000 & 1.1 & 50000 & 0.01 & 10000 & 0.857 & 0.875 \\
            v6 & 10000 & 1.1 & 10000 & 0.001 & 10000 & 0.862 & 0.877 \\
            v7 & 20000 & 1.1 & 20000 & 0.001 & 10000 & 0.862 & 0.877 \\
            v8 & 10000 & 1.1 & 10000 & 0.001 & 10000 & 0.856 & 0.876 \\
            v9 & 10000 & 1.05 & 10000 & 0.001 & 10000 & 0.854 & 0.869 \\
            v10 & 7000 & 1.1 & 10000 & 0.001 & 10000 & 0.854 & 0.877 \\
            v11 & 10000 & 1.1 & 2000 & 0.001 & 10000 & 0.857 & 0.876 \\
            v12 & 10000 & 1.1 & 10000 & 0.001 & 12000 & 0.855 & 0.877 \\
            \midrule
            Final & 10000 & 1.1 & 10000 & 0.01 & 12000 & 0.855 & 0.886 \\
            \bottomrule
        \end{tabular}
    \end{center}
\end{table}

\subsection{AG News}
\begin{figure}[htbp]
    \begin{center}
        \includegraphics[width=\linewidth]{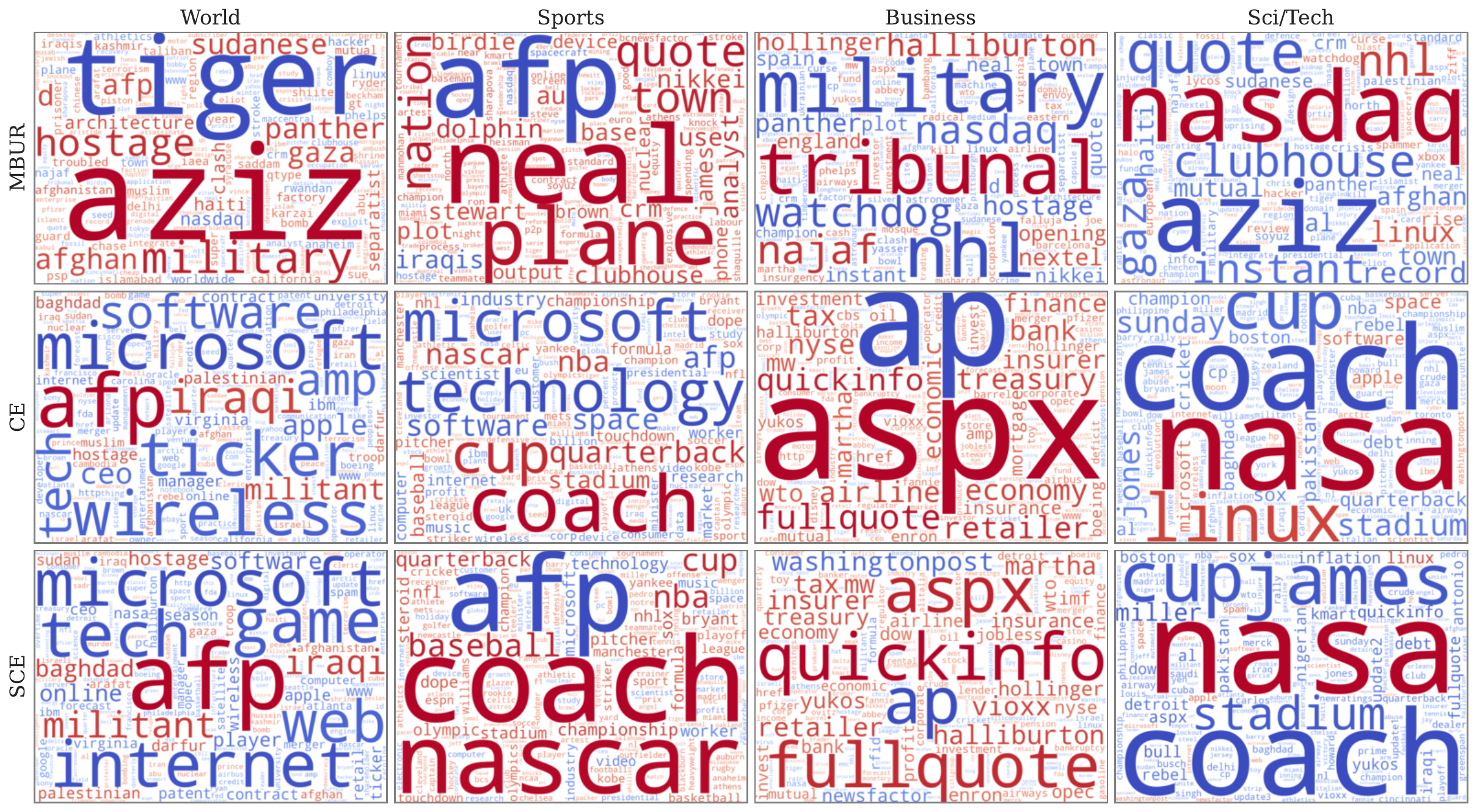}
    \end{center}
    \caption{Global interpretation for AG News across MBUR, CE, and SCE. The red color indicates positive word importance, while the blue color indicates negative importance. The relative magnitude of the importance is shown using the size of the word and the intensity of the color.}
    \label{fig:ag-news-wic-wordcloud}
\end{figure}

Figure~\ref{fig:ag-news-wic-wordcloud} shows the global interpretation for AG News using the same word-cloud procedure as IMDB, described in Section~\ref{sec:interpretability}.
MBUR finds a different set of top-magnitude words than CE and SCE, and for several classes these come from another class's vocabulary rather than the target class's own.
For Sci/Tech, MBUR's most important word is "nasdaq" (1.00), a Business term, while "nasa" (-0.59), the canonical Sci/Tech term shows negative importance. 
CE and SCE both correct determines "nasa" (1.00)  to be the most important and gives "nasdaq" near zero importance (0.06 and 0.02 respectively).
For Sports, MBUR gives high importance to "analyst", "device", "phone", and "crm" (0.79 to 0.90), all Business or Sci/Tech vocabulary, while "coach", the canonical Sports term is barely important (0.10). 
For the same class, CE and SCE correctly gives "coach" the maximum importance instead.
For Business, MBUR gives high positive magnitude to "najaf", "falluja", "mosque", and "insurgency" (0.66 to 0.90), while "economy" and "treasury" have negative importance scores.
CE and SCE again correct this, giving "economy" and "treasury" the highest importance instead.
Despite the accuracy gap between MBUR and Loss-Guided being relatively small on AG News, Loss-Guided model's interpretation is more richer and accurate.

\subsection{GoEmotions}

\begin{figure}[htbp]
    \begin{center}
        \includegraphics[width=\linewidth]{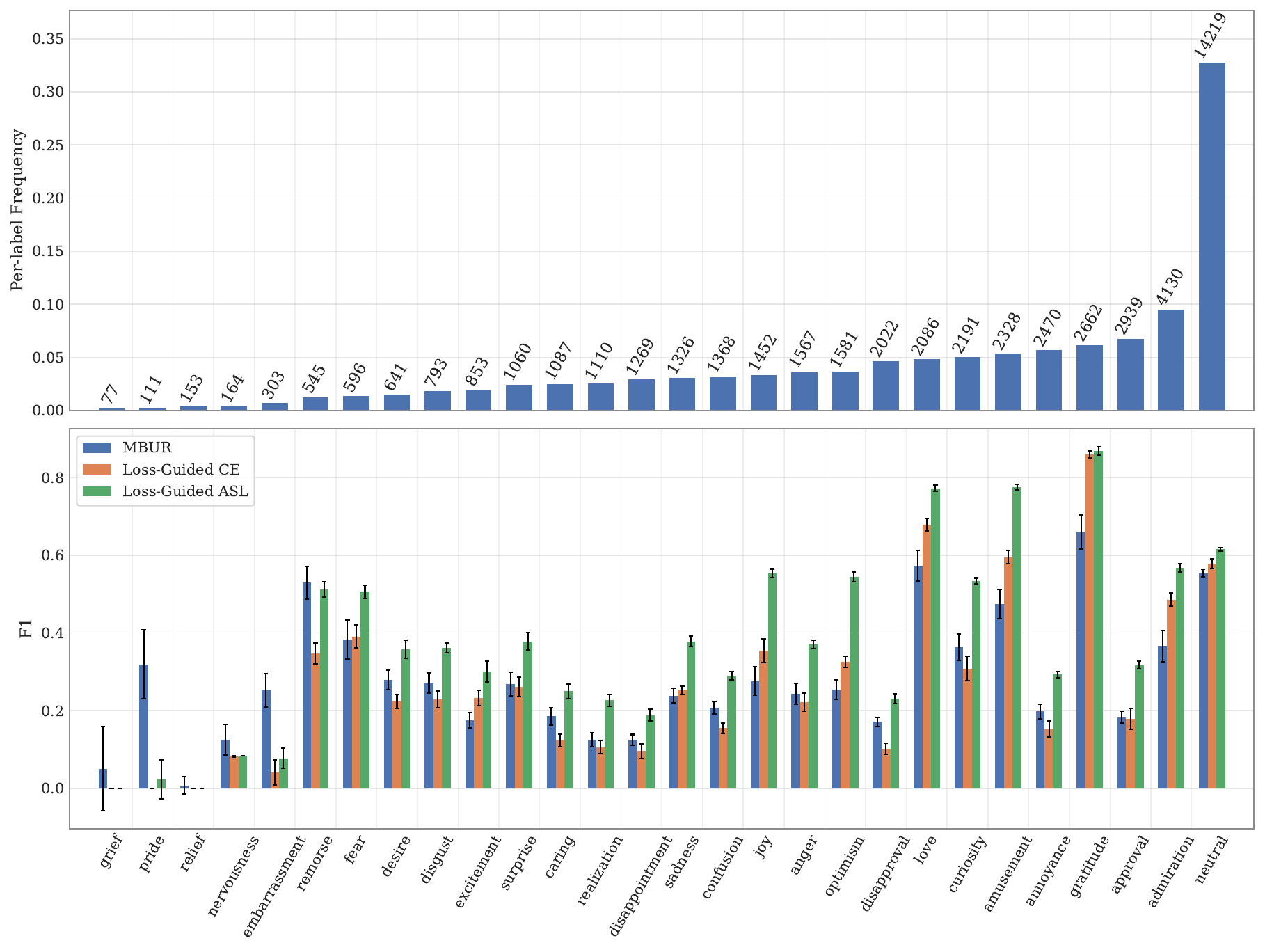}
    \end{center}
    \caption{GoEmotions label distribution and performance sorted by class rarity from rarest to most common. The top panel shows training set frequency per label. The bottom panel displays test F1 scores (mean $\pm$ std over 10 seeds) for MBUR, CE, and ASL.}
    \label{fig:go-emotions-per-class-f1}
\end{figure}

GoEmotions is a highly imbalanced multilabel dataset.
Figure~\ref{fig:go-emotions-per-class-f1} shows the per-label frequency in the training set alongside the per-class test F1 for MBUR, CE, and ASL, both sorted by class rarity.
Because the rarest classes have few training examples, none of the models perform reliably on them.
Loss-Guided consistently improves over MBUR on the moderate to common classes.

\begin{figure}[htbp]
    \begin{center}
        \includegraphics[width=\linewidth]{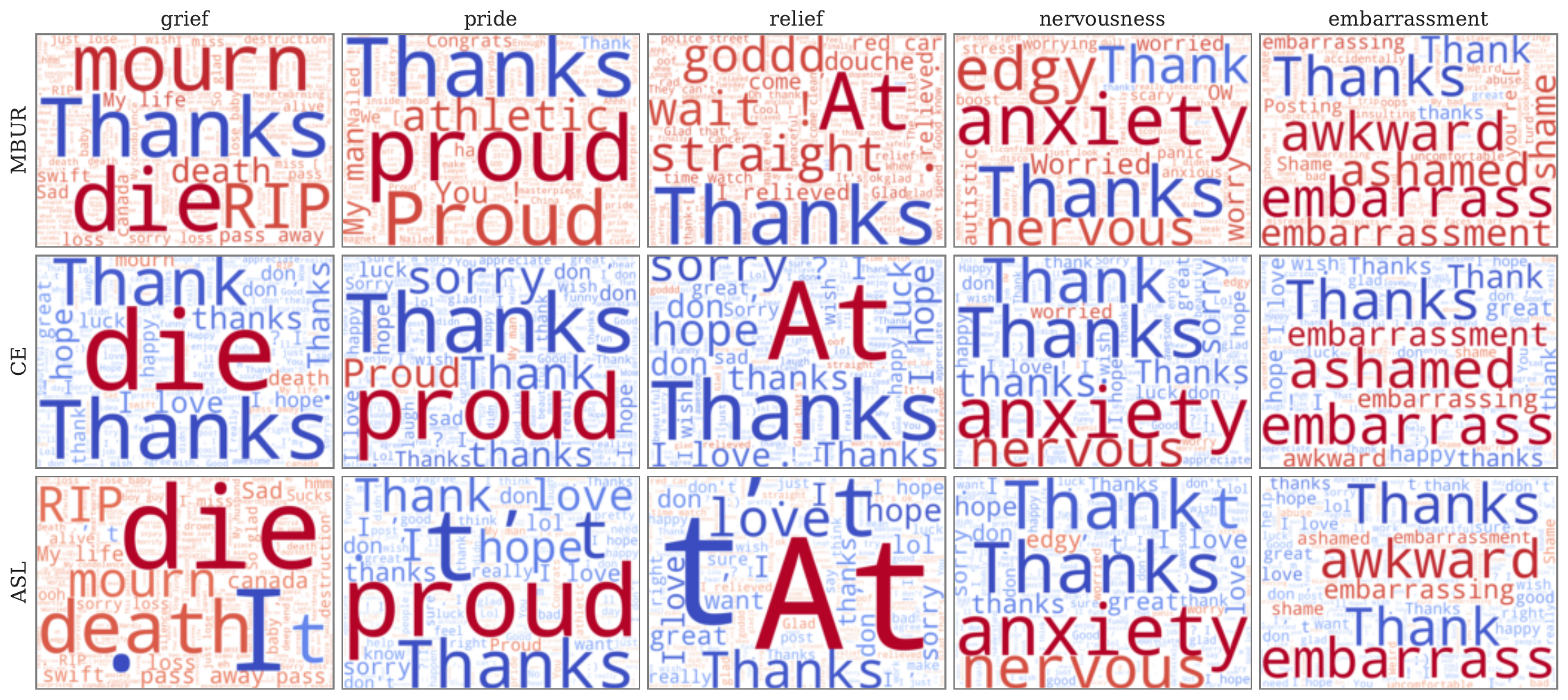}
    \end{center}
    \caption{Global interpretations for the five rarest GoEmotions classes.}
    \label{fig:go-emotions-wic-group1}
\end{figure}

\begin{figure}[htbp]
    \begin{center}
        \includegraphics[width=\linewidth]{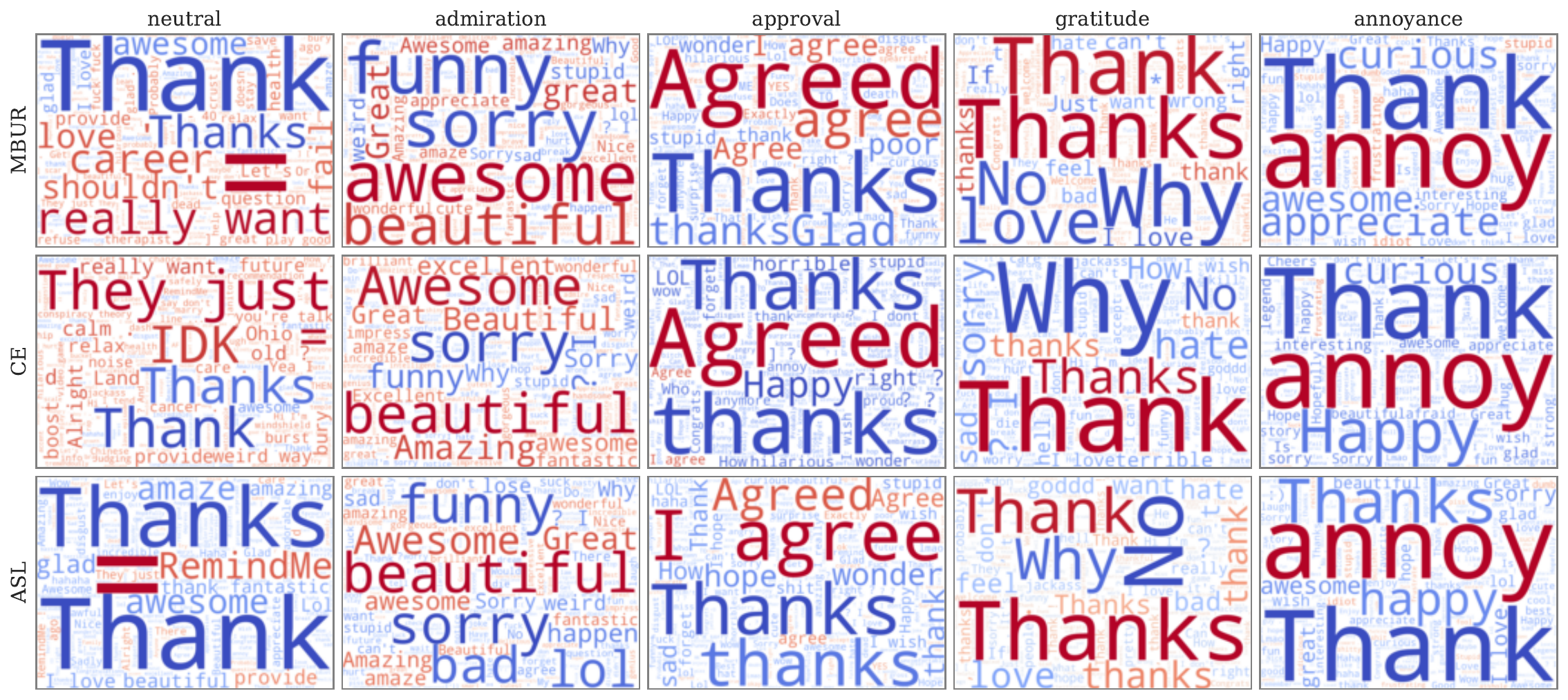}
    \end{center}
    \caption{Global interpretations for the five most common GoEmotions classes.}
    \label{fig:go-emotions-wic-group4}
\end{figure}

Figures~\ref{fig:go-emotions-wic-group1} and \ref{fig:go-emotions-wic-group4} show the global interpretability for the five rarest and five most common classes. Unlike IMDB and AG News, the three conditions largely agree here, each surfacing similar, topically coherent vocabulary per class.

For the common classes, this agreement lines up with equally strong F1 for all three conditions. Gratitude is a typical case, "Thanks" and "Thank" dominate under all three conditions. The one exception among the ten classes shown is neutral, a catch-all class with no consistent topic, where all three conditions surface generic, uninterpretable words instead.

For the rarest classes, the vocabulary agreement holds even where the predictive performance does not. On grief, Loss-Guided's test F1 is 0.00 for both CE and ASL, while MBUR reaches 0.05 with high variance. On pride, CE is at 0.00 and ASL at 0.02, while MBUR reaches 0.32. Despite this, "die", "mourn", and "RIP" dominate grief's global interpretation under all three conditions, and "proud", "Proud", and "athletic" dominate pride's under all three as well. The model has learned the right vocabulary for these two classes even where Loss-Guided never predicts them. Relief, the rarest class with only 153 positive training examples, is weaker across the board. All three conditions score near zero F1 there, and their top words mix genuine relief vocabulary such as "relieved" and "glad" with non-informative tokens such as "goddd" and "red car", under MBUR as much as under CE and ASL.

\subsection{CelebA}

\begin{figure}[htbp]
    \begin{center}
        \includegraphics[width=\linewidth]{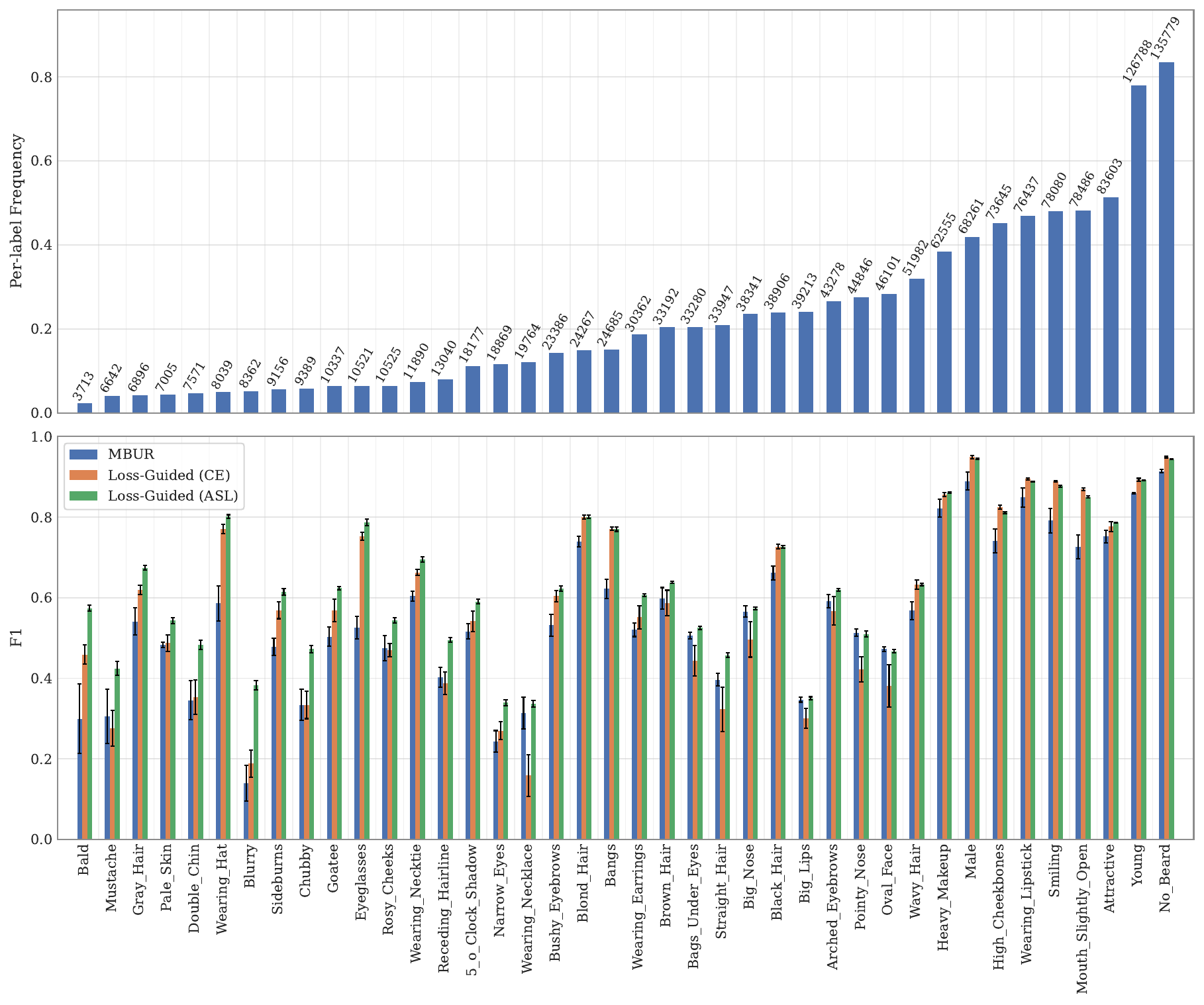}
    \end{center}
    \caption{(Top) per-label frequency in the CelebA training set. (Bottom) per-class test F1 for MBUR, CE, and ASL, mean $\pm$ std over 10 seeds. Both sorted by class rarity (rarest first).}
    \label{fig:celeba-per-class-f1}
\end{figure}

CelebA's 40 attributes are imbalanced as well, though less severely than GoEmotions (Figure~\ref{fig:celeba-per-class-f1}).
For ASL specific parameters, $\gamma_{+}=\{0, 2\}$ and $\gamma_{-}=\{2, 4, 6\}$ were tested.
$\gamma_{+}=2$ and $\gamma_{-}=4$ gave the best results and are thus reported in the paper.
Figure~\ref{fig:celeba-per-class-f1} shows the per-class test F1 for MBUR, CE, and ASL, sorted by class rarity.
ASL's gain over MBUR here is significantly larger on the rarest attributes.
The gains for more common labels are minimal.

\subsection{Computational Cost}
\label{sec:appendix-cost}

The runtime of TMs depends strongly on the implementation.
For the implementation used here, Table~\ref{tab:runtime} reports the mean runtime per seed for 100 training epochs on a single NVIDIA H100 GPU.
Loss-Guided TM is faster than MBUR on the larger datasets (AG News, GoEmotions, and CelebA), and slightly slower on the smaller ones.
Computing $\Delta L_k$ requires storing, for each clause, the vote sums and predictions obtained when the clause output is inverted, which needs $O(C \times M)$ additional memory.

\begin{table}[htbp]
    \caption{Mean runtime per seed for 100 training epochs on a single NVIDIA H100 GPU.}
    \label{tab:runtime}
    \begin{center}
        \begin{tabular}{lll}
            \toprule
            \textbf{Dataset} & \textbf{MBUR} & \textbf{Loss-Guided} \\
            \midrule
            MNIST & 5 min & 9 min (CE) \\
            FashionMNIST & 11 min & 14 min (CE) \\
            IMDB & 35 min & 54 min (CE) \\
            AG News & 77 min & 55 min (CE), 54 min (SCE) \\
            GoEmotions & 7.4 h & 2.5 h (CE), 2.3 h (ASL) \\
            CelebA & 8.4 h & 4.5 h (CE), 5.2 h (ASL) \\
            \bottomrule
        \end{tabular}
    \end{center}
\end{table}

\subsection{Post-hoc Calibration}
\label{sec:appendix-posthoc}

Post-hoc calibration methods such as temperature scaling \citep{guo2017CalibrationModernNeural} and Platt scaling \citep{platt1999ProbabilisticOutputsSupport} are commonly applied on top of trained models.
These methods can also be applied to MBUR and Loss-Guided TM to calibrate their probability scores.
To observe their effect, we applied temperature scaling to the multiclass and binary datasets and Platt scaling to the multilabel datasets.
Both methods are applied to the class sums, the unclipped $v_i$ (Eq.~\ref{eq:classic-vote-sum}) for MBUR and the normalized $z_i$ (Eq.~\ref{eq:class-sum}) for Loss-Guided TM.
The models are retrained with the configurations of Table~\ref{tab:hyperparameters}, holding out 10\% of the training set for validation, except for GoEmotions and CelebA, where the predefined validation splits are used.
The calibration parameters are fitted by minimizing the negative log-likelihood (NLL) on the validation set, and the ECE, NLL, and Brier score on the test set are reported in Table~\ref{tab:posthoc} (mean and std over 10 seeds).

With both models calibrated, Loss-Guided TM achieves the best ECE, NLL, and Brier score on five of the six datasets.
On MNIST, calibrated MBUR achieves slightly lower ECE, NLL, and Brier score, but the differences are practically negligible.
On IMDB, FashionMNIST, and AG News, CE without post-hoc calibration is better than calibrated MBUR on all three metrics.
Post-hoc calibration barely changes the CE models on the multiclass and binary datasets, indicating that CE training already yields calibrated probabilities.
Post-hoc calibration is therefore complementary to the loss-guided learning mechanism.

\begin{table}[htbp]
    \caption{ECE, NLL, and Brier score on the test set without (Uncalibrated) and with (Post-hoc) post-hoc calibration, mean $\pm$ std over 10 seeds. The best value per column and dataset is in bold.}
    \label{tab:posthoc}
    \begin{center}
        \resizebox{\linewidth}{!}{%
        \begin{tabular}{llcccccc}
            \toprule
            & & \multicolumn{2}{c}{\textbf{ECE}} & \multicolumn{2}{c}{\textbf{NLL}} & \multicolumn{2}{c}{\textbf{Brier}} \\
            \cmidrule(lr){3-4} \cmidrule(lr){5-6} \cmidrule(lr){7-8}
            \textbf{Dataset} & \textbf{Model} & Uncalibrated & Post-hoc & Uncalibrated & Post-hoc & Uncalibrated & Post-hoc \\
            \midrule
            \multirow{2}{*}{MNIST} & MBUR & 0.0056$\pm$0.0002 & \textbf{0.0016$\pm$0.0001} & 0.0827$\pm$0.0027 & \textbf{0.0397$\pm$0.0017} & 0.0337$\pm$0.0008 & \textbf{0.0193$\pm$0.0011} \\
             & CE & \textbf{0.0015$\pm$0.0001} & 0.0017$\pm$0.0001 & \textbf{0.0466$\pm$0.0026} & 0.0467$\pm$0.0023 & \textbf{0.0230$\pm$0.0011} & 0.0230$\pm$0.0010 \\
            \midrule
            \multirow{2}{*}{FashionMNIST} & MBUR & 0.0255$\pm$0.0015 & 0.0055$\pm$0.0003 & 0.4166$\pm$0.0076 & 0.2854$\pm$0.0030 & 0.1767$\pm$0.0030 & 0.1438$\pm$0.0017 \\
             & CE & \textbf{0.0045$\pm$0.0004} & \textbf{0.0045$\pm$0.0004} & \textbf{0.2727$\pm$0.0030} & \textbf{0.2724$\pm$0.0030} & \textbf{0.1409$\pm$0.0020} & \textbf{0.1409$\pm$0.0020} \\
            \midrule
            \multirow{2}{*}{IMDB} & MBUR & 0.0902$\pm$0.0009 & 0.0179$\pm$0.0031 & 1.2310$\pm$0.0136 & 0.3488$\pm$0.0018 & 0.1189$\pm$0.0006 & 0.1068$\pm$0.0006 \\
             & CE & \textbf{0.0154$\pm$0.0015} & \textbf{0.0120$\pm$0.0021} & \textbf{0.2832$\pm$0.0011} & \textbf{0.2827$\pm$0.0011} & \textbf{0.0853$\pm$0.0003} & \textbf{0.0850$\pm$0.0003} \\
            \midrule
            \multirow{3}{*}{AG News} & MBUR & 0.0654$\pm$0.0036 & 0.0200$\pm$0.0042 & 0.6014$\pm$0.0335 & 0.3688$\pm$0.0095 & 0.2099$\pm$0.0047 & 0.1785$\pm$0.0052 \\
             & CE & \textbf{0.0112$\pm$0.0009} & \textbf{0.0111$\pm$0.0007} & \textbf{0.3231$\pm$0.0020} & \textbf{0.3232$\pm$0.0020} & 0.1600$\pm$0.0012 & 0.1598$\pm$0.0012 \\
             & SCE & 0.0259$\pm$0.0005 & 0.0136$\pm$0.0007 & 0.3991$\pm$0.0019 & 0.3281$\pm$0.0008 & \textbf{0.1589$\pm$0.0011} & \textbf{0.1557$\pm$0.0007} \\
            \midrule
            \multirow{3}{*}{GoEmotions} & MBUR & 0.1577$\pm$0.0078 & 0.0078$\pm$0.0005 & 0.5131$\pm$0.0178 & 0.1129$\pm$0.0006 & 0.0948$\pm$0.0041 & 0.0292$\pm$0.0001 \\
             & CE & \textbf{0.0141$\pm$0.0005} & \textbf{0.0070$\pm$0.0004} & \textbf{0.1178$\pm$0.0005} & 0.1105$\pm$0.0003 & \textbf{0.0291$\pm$0.0001} & 0.0284$\pm$0.0001 \\
             & ASL & 0.2392$\pm$0.0012 & 0.0094$\pm$0.0001 & 0.3403$\pm$0.0016 & \textbf{0.1064$\pm$0.0001} & 0.0923$\pm$0.0006 & \textbf{0.0273$\pm$0.0001} \\
            \midrule
            \multirow{3}{*}{CelebA} & MBUR & 0.1533$\pm$0.0070 & 0.0250$\pm$0.0003 & 0.4841$\pm$0.0208 & 0.2962$\pm$0.0009 & 0.1301$\pm$0.0034 & 0.0927$\pm$0.0003 \\
             & CE & \textbf{0.0285$\pm$0.0014} & \textbf{0.0215$\pm$0.0002} & \textbf{0.2566$\pm$0.0010} & \textbf{0.2524$\pm$0.0003} & \textbf{0.0795$\pm$0.0003} & \textbf{0.0787$\pm$0.0001} \\
             & ASL & 0.1803$\pm$0.0007 & 0.0217$\pm$0.0002 & 0.3882$\pm$0.0007 & 0.2541$\pm$0.0002 & 0.1171$\pm$0.0002 & 0.0793$\pm$0.0001 \\
            \bottomrule
        \end{tabular}%
        }
    \end{center}
\end{table}

\subsection{Ablation of Clause Feedback and Weight Updates}
\label{sec:appendix-ablation}

Loss-Guided TM differs from MBUR in both the clause feedback and the weight updates.
Both are derived from the same explicit loss, the clause feedback from the change in loss when a clause output is inverted and the weight updates from its gradient.
Thus, they form a single learning mechanism rather than two independent modifications of MBUR.
To examine whether the two parts can be separated, we experiment with two ablations and study their effect.
Both ablations use the configurations of Table~\ref{tab:hyperparameters} and are evaluated over 5 seeds.
Table~\ref{tab:ablation} compares both ablations with MBUR and Loss-Guided TM (CE).

\begin{table}[htbp]
\caption{Test accuracy and ECE of the two ablations, compared with MBUR and Loss-Guided TM (CE), mean $\pm$ std over 5 seeds. Ablation A uses loss-guided feedback with integer weight updates, and Ablation B uses MBUR feedback with gradient-trained weights. (\ref{sec:appendix-ablation})}
    \label{tab:ablation}
    \begin{center}
        \begin{tabular}{llcc}
            \toprule
            \textbf{Dataset} & \textbf{Model} & \textbf{Accuracy (\%)} & \textbf{ECE} \\
            \midrule
            \multirow{4}{*}{MNIST} & MBUR & \textbf{98.66 $\pm$ 0.08} & 0.0057 $\pm$ 0.0002 \\
              & Loss-Guided (CE) & 98.42 $\pm$ 0.12 & \textbf{0.0015 $\pm$ 0.0001} \\
              & Ablation A & 97.50 $\pm$ 0.21 & 0.0047 $\pm$ 0.0004 \\
              & Ablation B & 85.55 $\pm$ 0.72 & 0.3204 $\pm$ 0.0023 \\
            \midrule
            \multirow{4}{*}{FashionMNIST} & MBUR & 90.39 $\pm$ 0.11 & 0.0261 $\pm$ 0.0010 \\
              & Loss-Guided (CE) & \textbf{90.50 $\pm$ 0.27} & \textbf{0.0046 $\pm$ 0.0003} \\
              & Ablation A & 83.10 $\pm$ 1.22 & 0.0312 $\pm$ 0.0025 \\
              & Ablation B & 71.22 $\pm$ 1.42 & 0.3174 $\pm$ 0.0022 \\
            \midrule
            \multirow{4}{*}{IMDB} & MBUR & 85.45 $\pm$ 0.12 & 0.0942 $\pm$ 0.0019 \\
              & Loss-Guided (CE) & \textbf{88.54 $\pm$ 0.12} & \textbf{0.0136 $\pm$ 0.0015} \\
              & Ablation A & 83.51 $\pm$ 0.41 & 0.0967 $\pm$ 0.0042 \\
              & Ablation B & 85.11 $\pm$ 0.06 & 0.0936 $\pm$ 0.0017 \\
            \midrule
            \multirow{4}{*}{AG News} & MBUR & 88.79 $\pm$ 0.25 & 0.0628 $\pm$ 0.0013 \\
              & Loss-Guided (CE) & \textbf{89.79 $\pm$ 0.17} & \textbf{0.0110 $\pm$ 0.0009} \\
              & Ablation A & 88.53 $\pm$ 0.43 & 0.0565 $\pm$ 0.0020 \\
              & Ablation B & 88.88 $\pm$ 0.12 & 0.3373 $\pm$ 0.0005 \\
            \bottomrule
        \end{tabular}
    \end{center}
\end{table}

\begin{enumerate}
    \item[A.] \textbf{Loss-Guided Feedback with Integer Weights.}
        In this ablation, the Loss-Guided clause feedback is used with the integer weight updates of MBUR.
        The weights change in steps of $\pm 1$ only for clauses receiving Type Ia or Type II feedback, similar to MBUR.
        The accuracy of Ablation A drops below that of MBUR on MNIST, FashionMNIST, and IMDB, and is on par with MBUR on AG News.
        The ECE for Ablation A is also worse compared to Loss-Guided TM, thereby losing the calibration improvements.
        On GoEmotions, training of Ablation A diverges with both CE and ASL, the training loss increasing until learning stops completely with a macro F1 of 0.

    \item[B.] \textbf{MBUR Feedback with Gradient-Trained Weights.}
        In this ablation, the MBUR clause feedback is used with the gradient-based weight updates of Loss-Guided TM, trained on the CE loss.
        The values of $T$ and $q$ for MBUR and of $\eta$ for Loss-Guided TM are taken from Table~\ref{tab:hyperparameters}.
        The accuracy of Ablation B is on par with MBUR on IMDB and AG News, and far below MBUR on MNIST and FashionMNIST.
        The ECE is much worse than that of both MBUR and Loss-Guided TM, except on IMDB, where it is close to that of MBUR.
        On GoEmotions, Ablation B does not learn, with a macro F1 below 3\% compared with 28.9\% for MBUR.

\end{enumerate}

Neither ablation recovers the accuracy or the calibration of Loss-Guided TM, and both fail on the multilabel GoEmotions dataset.
With the hyperparameters of Table~\ref{tab:hyperparameters}, the loss-guided clause feedback and the gradient-based weight updates are therefore effective only when used together, as parts of the same learning mechanism.

\end{document}